\documentclass[letterpaper, 10 pt, conference]{ieeeconf}

\IEEEoverridecommandlockouts                              

\usepackage{graphicx} 
\usepackage{epsfig} 
\usepackage{mathptmx} 
\usepackage{amsmath} 
\usepackage{amssymb}  
\usepackage{xcolor}
\usepackage[hidelinks]{hyperref}
\usepackage{url}
\usepackage{soul}

\usepackage{mathrsfs}
\usepackage{subcaption}
\usepackage{float}
\usepackage[11pt]{moresize}
\usepackage[english]{babel}
\usepackage[utf8]{inputenc}
\usepackage{csquotes}
\usepackage{makecell}

\usepackage{colortbl}

\begin{document}
%
\title{\LARGE \bf Visual-based Safe Landing for UAVs in Populated Areas: Real-time Validation in Virtual Environments}
%
%
%

\author{Héctor Tovanche-Picón$^1$, Javier González-Trejo$^2$, Ángel Flores-Abad$^3$ and Diego Mercado-Ravell$^{2,4}$
\thanks{$^1$H. Tovanche-Picón is with Universidad Autónoma de Ciudad Juárez UACJ, Mexico. $^2$J. González-Trejo and D. Mercado-Ravell are with the Center for Research in Mathematics CIMAT AC, Mexico. $^3$A. Flores-Abad is with University of Texas at El Paso. 
$^4$D. Mercado-Ravell is also with Investigadores CONACyT 
(email: {\tt\footnotesize hector.tovanche@uacj.mx; javier.gonzalez@cimat.mx; afloresabad@utep.edu; diego.mercado@cimat.mx)}}
}

\maketitle
\thispagestyle{empty}
\pagestyle{empty}
\begin{abstract}
Safe autonomous landing for Unmanned Aerial Vehicles (UAVs) in populated areas is a crucial aspect for successful urban deployment, particularly in emergency landing situations. Nonetheless, validating autonomous landing in real scenarios is a challenging task involving a high risk of injuring people. In this work, we propose a framework for real-time safe and thorough evaluation of vision-based autonomous landing in populated scenarios, using photo-realistic virtual environments. We propose to use the Unreal graphics engine coupled with the AirSim plugin for drone´s simulation, and evaluate autonomous landing strategies based on visual detection of Safe Landing Zones (SLZ) in populated scenarios. Then, we study two different criteria for selecting the ``best" SLZ, and evaluate them during autonomous landing of a virtual drone in different scenarios and conditions, under different distributions of people in urban scenes, including moving people. We evaluate different metrics to quantify the performance of the landing strategies, establishing a baseline for comparison with future works in this challenging task, and analyze them through an important number of randomized iterations. The study suggests that the use of the autonomous landing algorithms considerably helps to prevent accidents involving humans, which may allow to unleash the full potential of drones in urban environments near to people.
\end{abstract}


%
\IEEEpeerreviewmaketitle

\section{Introduction}
The role that Unmanned Aerial Vehicles (UAVs) have adopted in recent years in diverse application fields such as precision agriculture, package delivery, recreational activities, among others, have been greatly improved thanks to the high mobility and flexibility that this kind of devices provides jointly with the increasingly lower manufacturing cost. However, even with the enormous progress achieved with these vehicles, their true potential has been hindered in urban scenarios, mainly due to the risk of injuring people in case an accident, considerably limiting their applications to controlled environments or rural areas. Accordingly, these vehicles require to fulfill several safety constrains and legal regulations \cite{FAA2021}, for instance, in the USA the regulations include pilot certification and operational limitations such that the vehicles must weight less than 55 pounds, they cannot fly above 400 feet above the sea or the top of structures, even more, one of the harshest limitations forbids drones to fly over people, preventing the use of UAVs in many interesting civilian applications, and even so, this cannot guarantee the safety of people in case of emergency landings caused by loss of signal, adverse environmental conditions, low battery or human errors. In that sense, providing UAVs with emergency landing protocols, including autonomous landing, will help to prevent accidents, considerably boosting drone's potential for civilian applications in urban areas, close to people. 

The autonomous landing capability has been previously discussed by the scientific community, mainly focusing on detection of predefined landing tags \cite{garcia2002towards}, or finding flat landing surfaces, free of obstacles \cite{johnsonil_vision_guided_landin_auton_helic_hazar_terrain}. However, autonomous safe landing in populated scenarios is a much more recent and very challenging task. Only a few related works are to be found in the recent literature, mainly using Deep Neural Networks (DNN) for people detection from UAV's images \cite{tzelepi19_graph_embed_convol_neural_networ, castellano20_crowd_detec_drone_safe_landin}. However, autonomous landing in populated environments is still an almost unexplored problem.


\begin{figure}[t]
\centerline{\includegraphics[width=0.39\textwidth]{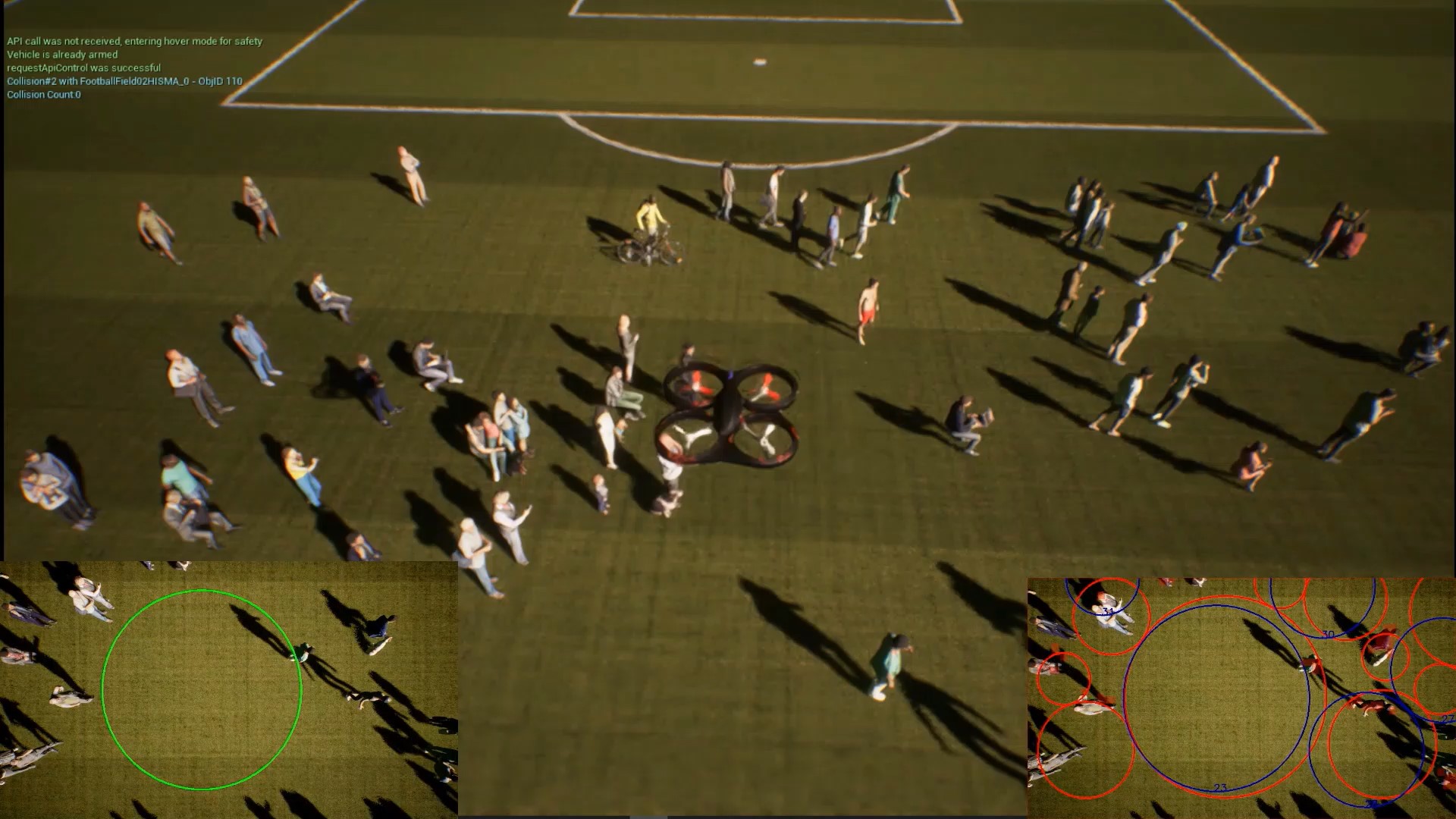}}
\caption{Validation of UAVs autonomous landing in populated areas using virtual environments.}
\label{fig:landingexample}
\end{figure}

Before deployment of autonomous landing strategies in populated environments, it is necessary to test them numerous times in order to quantify their performance and reliability, specially in conditions where people is present (see Fig. \ref{fig:landingexample}). For an algorithm that looks to avoid injuries and prevent accidents it is a paradox to be tested in real conditions before a proper validation, since the system is likely to put in danger the same people that it is aiming to protect. It is also very important to widely test beforehand these vision-based algorithms to characterize its performance in the desired conditions, and compare different approaches to fine-tune them and select the best suited ones. In this regard the virtual environments provide a safe and inexpensive manner to test algorithms for autonomous vehicles. The virtual environments must offer close-to-real-world conditions to provide a useful approximation of the performance of the tested system. Also, the physics and dynamics of autonomous vehicles can be replicated to have a closer to real world behavior. However, despite the big efforts in generating realistic simulations, there will always exist a domain gap between simulated - real world scenarios to take into consideration \cite{Sankaranarayanan2017}. In this sense, the use of virtual environments for testing autonomous navigation cannot guarantee the success in real-world applications, but provides a powerful safe and cheap way, maybe the only available one, to test and compare these algorithms as a first validation step.

In previous work \cite{gonzalez-trejo21_light_densit_map_archit_uavs}, we have proposed a lightweight DNN architecture for crowds detection, suitable for real-time embedded applications, and provided a first solution to detect Safe Landing Zones (SLZ) in populated environments, later on, in \cite{gonzalez-trejo21_visual_based_safe_landin_uavs_popul_areas}, we considerably improved the SLZ detection, considering the camera movement and tracking multiple instances of the SLZ along time. The performance of the strategy was evaluated offline using two public image data-sets with aerial views, Venice \cite{Liu_2019_CVPR} and VisDrone \cite{zhu2020vision}, as well as with data taken from a drone manually operated during landing missions in real populated scenarios. Nevertheless, due to the harsh safety constrains involved in the experiments, real-time validation during fully autonomous landing missions is still an important open challenge before deployment in real missions. 

Henceforth, in this paper we present a pipeline using virtual environments to evaluate autonomous landing algorithms (see Figure \ref{fig:landingexample}), study their performance under different conditions and compare different solutions. The proposed solution must select in real-time a target SLZ according to different criteria, and command a simulated UAV to move to a suitable region and perform autonomous landing without putting people at risk. The drone and its physics are simulated using the Airsim plug-in. The virtual environment was rendered using Unreal Engine 4, a powerful graphics engine that offers photo-realistic characteristics, providing a close representation of urban scenarios with people. We tested the performance of the algorithm for autonomous landing in several randomized and dynamic environments, unknown to the UAV, with people present in the scene. We performed a wide range of tests under different conditions and scenarios, allowing some of the people to move with random walks, using diverse domain randomization such as density (people by square meter), people's initial position and appearance,  floor texture, and weather conditions. The study suggests that the use of visual-based autonomous landing strategies in populated environments could considerably help to reduce the risk of accidents involving humans. Particularly, we have found that the use of the older SLZ as selection criteria tends to be more robust to the movement of the people in the scene. Moreover, the proposed framework offers a baseline for comparison of different strategies, allowing us to test them qualitatively and fine-tune them before real deployment. 

The main contributions are summarized as follows:
\begin{enumerate}
    \item Real-time implementation of an autonomous landing algorithm based in the SLZ detection algorithm proposed in \cite{gonzalez-trejo21_visual_based_safe_landin_uavs_popul_areas}, in a photo-realistic environment using a simulated Drone. 
    \item Qualitative validation of the proposed pipeline trough multiple iterations using domain randomization techniques in key aspects such as the people characteristics, people number and initial position, people´s movements, texture and background of the scene, illumination changes, weather conditions, etc.
    \item A safe environment to test vision-based algorithm without putting people at risk and reducing the time and resources required for testing in real-word.
    \item A baseline framework for safe and accurate comparison with other future works aiming to solve the problem of autonomous landing in populated environments.
\end{enumerate}

The rest of this work is organized as follows: In Section \ref{sec2} we discuss the SLZ detection algorithm. In Section \ref{sec3} we introduce the validation framework. Section \ref{sec4} presents the main results obtained in the experiments. Finally, in Section \ref{sec5} we provide the conclusions.

\section{SLZ algorithm} \label{sec2}


The main objective of the SLZ detection algorithm is to propose zones free of people, in the means of landing a UAV without crashing it against the crowd. The algorithm proposes SLZs in real-world coordinates during the mission duration, accounting for the camera and crowd movements \cite{gonzalez-trejo21_visual_based_safe_landin_uavs_popul_areas}. For that matter, the algorithm is divided into three sub modules, 1) a lightweight DNN density map generator that separates the people in the image, 2) a custom algorithm to find suitable circular SLZs in real-world coordinates from the people-free regions, and 3) a multiple-instances object trackers by means of Kalman Filters (KF) that tracks the SLZs between each crowd segmentation, adding robustness for the sudden changes in the segmentation due to the camera and people´s movement (note that in real world people moves at will with unknown dynamics).

\subsection{Crowd detection}

To obtain suitable SLZ candidates, the first step of our algorithm is to infer the crowd´s location from video streams in real-time. Most of the methods in the literature, and the one used by our algorithm, use density maps, which convey the spatial location of crowds with their people count in an image. These density maps are trained to detect the people's heads in the image, given that the head is the most visible part of the people from aerial views, providing superior performance in scenarios where the crowd density is high, and the visibility of the people's bodies is reduced due to occlusions. In such scenarios, the landing of the UAV in SLZ is crucial. In that regard, we developed a lightweight neural network called Bayes Loss Pruned Compact Convolutional Neural Network (BL Pruned CCNN) to generate these density maps at low computational cost, suited for accurate people's localization in embedded devices \cite{gonzalez-trejo21_light_densit_map_archit_uavs}.

The DNN is composed of $9$ fully convolutional layers divided by a head and a backbone, following the Multi-Column Convolutional Neural Network (MCNN) architecture described in \cite{zhang16_singl_image_crowd_count_multi}. The head is constituted by three columns of one layer, each with different kernel sizes that capture features of small, medium, and big-sized heads in the image. At the end of the first layer, the backbone process the features obtained by the head to generate the density map, it is a heat map encoding the inferred location and distribution of the people in the scene. We trained and fine-tuned the neural network using the Bayes Loss \cite{ma2019bayesian} to achieve a Mean Square Error (MSE) of $241.77$ in the crowd counting task. Finally, we reduced the original number of parameters from $0.075$ M to $0.065$ by pruning its channels, and successfully implemented it in real-time on an embedded processor.

\subsection{Safe Landing Zone Proposals}
Once our density map generator infers the crowd location, the next step is to find the SLZ in real-world coordinates. From the output of the density map $D$, we obtain a binary occupancy map $O$ with pixels values equal to either $0$ or $255$, representing crowded or free zones, respectively. The pixel values from $D$ are normalized by applying a threshold $\min\{D_{i, j}\}$, where ${i, j}$ are pixel coordinates. For safety reasons, a dilation transformation is applied to the occupancy map $O$, to overestimate the crowd location.

From there, using the camera pose info from the propriosceptive sensors in the drone, we project $O$ from the image plane to the world frame. Henceforth, we model our projection transformation using the thin lens camera model, where we define three reference frames. $\mathbf{F_W}$ represents the inertial or world reference. The UAV frame, located at its mass center, is referred as $\mathbf{F_B}$. The homogeneous transformation to project coordinates from $\mathbf{F_B}$ to $\mathbf{F_W}$ is defined as $\mathbf{T}^W_B \in SE(3)$, and is obtained from the on-board UAV sensor. The camera frame is referred as $\mathbf{F_C}$, with its transformation from $\mathbf{F_B}$ to $\mathbf{F_C}$ defined as $\mathbf{T}^B_C \in SE(3)$, which is known a priori.

An important element of our pipeline that allows us to project from the image to the inertial frame, is the assumption on which all the heads are approximately located in a common plane known as the head's plane $P$ in $\mathbf{F_W}$, which is parallel to the floor at the average people's height $h_t$. 

Having our model defined, to perform the projection of the occupancy map $O$ to the head's plane $P$, denoted as $O^W$, we define a grid $G \in P$, as proposed in the sampler module of the Spatial Transformer Network \cite{jaderberg2015spatial}. Each element $G_{i, j}$ stores coordinates $\{x_W, y_W\}$ of the plane $P$, referred as $(G_{i, j}^x, G_{i, j}^y)$. For each $G_{i, j}$, we map a coordinate $(x_I, y_I)$ in the image $I$ as follows:

\begin{equation}
\label{eq:projection}
   \lambda \cdot (x_I, y_I, 1)^T = \mathbf{K}  \mathbf{T}^C_B \mathbf{T}^B_W \cdot (G^{x}_{i,j},G^{y}_{i,j}, h_h, 1)^T. 
\end{equation}
where $\lambda$ is a scale factor and $\mathbf{K}$ are the intrinsic parameters of the camera, defined in the Unreal Engine. Using Eq. \ref{eq:projection}, we query the value for each coordinate of the occupancy map $O^W$.

Once we have our occupancy map in the head's plane, we obtain a distance map $C$ from where each nonzero (people-free zone) pixel contains the distance to the nearest zero pixel value (crowded zone). By calculating the $\max\{C\}$, we effectively find the radius and center of the biggest circular SLZ. This SLZ is drawn into the occupation map $O^W$, repeating the process up to $N_p$ times.
\subsection{Multiple Landing Zone Tracking}
By themselves, the $N_p$ SLZ found can be used to flag the location in the head's plane where the UAV can land, nevertheless, due to the constant movement of both the camera and the crowd through the scenario, the location of the SLZ may vary abruptly, potentially pointing the UAV to a landing zone with people on it. In that regard, we track the location of multiple SLZ using KFs, smoothing out the movement of the SLZ between each $k$ frame of the video stream, ensuring their temporal consistency and adding robustness against people moving. For each SLZ $i$, let us consider a vector state at time $k$ equal to
\begin{equation}
\mathbf{x}_{k}^{(i)} =  \left[ x_k^{(i)},y_k^{(i)},r_k^{(i)},\Dot{x}_k^{(i)},\Dot{y}_k^{(i)}, \Dot{r}_k^{(i)} \right]^T,    
\end{equation}
where $(x_k^{(i)}, y_k^{(i)})$ are the coordinates of the $i$-th SLZ in the plane $P$, $r_k^{(i)}$ the radius, and $\Dot{x}_k^{(i)},\Dot{y}_k^{(i)}, \Dot{r}_k^{(i)}$ the rates of change of the center coordinates and the radius, respectably. We model the displacement of the SLZ due to the movement of both the camera and the crowd using the constant velocity model, accounting for the acceleration in the process noise $\omega$ with normal distribution \cite{saho18_kalman_filter_movin_objec_track}, it is
\begin{equation}
    \mathbf{x}_{k+1}^{(i)} = 
    \begin{bmatrix}
        \mathbf{I}_3 & \Delta t \mathbf{I}_3\\
        \mathbf{0} & \mathbf{I}_3
    \end{bmatrix}
    \mathbf{x}_{k}^{(i)}+
 \mathbf{\omega}\ \ \ |\ \ \ \mathbf{\omega}\sim \mathcal{N}(\mathbf{0},\,\mathbf{Q})    ,
\end{equation}
where $ \mathbf{0}$ is a matrix of zeros, $\mathbf{I}_3 \in \mathbb{R}^{3 \times 3}$ is the identity matrix, $\Delta t$ is the time increment. The process noise covariance, which models the acceleration is defined as
\begin{equation}
 \mathbf{Q}=\sigma_a \begin{bmatrix}
 \frac{\Delta t^4}{4} \mathbf{I}_3& \frac{\Delta t^3}{2} \mathbf{I}_3\\
\frac{\Delta t^3}{2} \mathbf{I}_3& \Delta t^2 \mathbf{I}_3
\end{bmatrix} ,
\end{equation}
where $\sigma_a$ represents the acceleration uncertainty, obtained empirically. We use the SLZ measurements as they are inferred by the previous step. More formally, the measurement model is defined as
\begin{equation}
    \mathbf{z}^{(i)}_k = 
    \begin{bmatrix}
        \mathbf{I}_3 & \mathbf{0}
    \end{bmatrix}
    \mathbf{x}^{(i)}_k + \mathbf{\nu}\ \ \ |\ \ \ \mathbf{\nu}\sim \mathcal{N}(\mathbf{0},\,\mathbf{R})    ,
\end{equation}
where $\mathbf{\nu}$ is the measurement noise with covariance matrix $R$.

The algorithm maintains a $M_p$ number of KF trackers where $M_p < N_p$. We associate a KF to a SLZ candidate through the Intersection over Union (IoU) between the area of the detected SLZ $A_i$ and the SLZ tracked by the KF $A_j$, it is
\begin{equation}
    IoU = \frac{A_i \cap A_j}{A_i \cup A_j}.
    \label{eq:iou}
\end{equation}
Having calculated the IoU for all pairs of candidates and KFs, we associate a candidate with the KF that exceeds a threshold $\mu$ using the Hungarian algorithm \cite{bruff2005assignment}. Each time a KF has no SLZ associated, a counter is increased. If the counter exceeds a threshold $\mu_1$, the tracker is eliminated. Finally, since for some frames SLZ can not be calculated, we can use the prediction step of the KF until the next batch of SLZ are inferred. 

\section{Evaluation Framework} \label{sec3}
\subsection{Virtual Environment}
The experiments were performed with a I7 6700 processor, 32 GBs RAM memory, and due to the heavy computational burden to run both the virtual environment and the autonomous landing algorithm in real-time, a RTX 2080 graphics card from Nvidia was used. 

As presented in \cite{Shah2018}, the Airsim plug-in provides a high fidelity simulation for UAVs, providing a full integration between a flight controller, physics engine, inertial sensors, video cameras, depth sensors and another capabilities. Airsim can be integrated with Unreal Engine and Unity, two of the most used engines for 3D recreation. In this work the elected engine is Unreal Engine 4 due to its photo-realistic rendering capabilities. 

\begin{figure}[t]
\centerline{\includegraphics[width=0.43\textwidth]{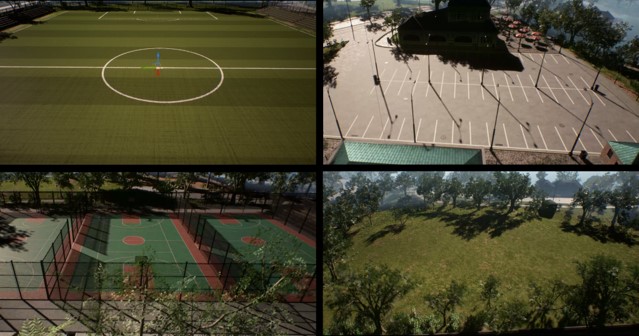}}
\caption{Sample scenarios from the CityPark Environment \cite{SilverTm2021}, containing a soccer field, parking lot, basketball courts and field with trees.}
\label{fig:airsimenvironments}
\end{figure}

In order to recreate realistic urban scenarios, the open source package City Park Environment collection Lite was used \cite{SilverTm2021}, it contains high quality assets ideal for recreating city parks. The scenarios present in this environment are suitable to test our algorithms providing a realistic quality and a rich textures condition, in a wide variety of urban-like environments. Figure \ref{fig:airsimenvironments} presents some samples of the selected scenarios for this work, including a soccer field, a parking lot, basketball courts and a field with trees. 

To represent crowds in the virtual environment, the \textit{Twinmotion} posed Human pack was selected, this pack contains $142$ 3D scanned different characters with high resolution textures, considering diverse characteristics such as age, gender, racial diversity, and a wide range of clothes, head accessories and poses. This ``actors´´ (in Unreal 4 context) could be identified by the algorithm as crowds allowing the cross domain application of a model trained with real-world data and deployed in a virtual environment, allowing us to test the visual-based autonomous landing algorithms without putting people at any risk. 
\subsection{Domain randomization}
The main purpose of domain randomization in a validation stage of an algorithm is to provide enough simulated variability to represent real-world conditions. We randomized from uniform distributions aspects of each of this domains in this work: landing scenario; people characteristics (gender, racial group, clothes, accessories, etc.); number of people, initial position and pose; actors movement; texture of the floor; scene illumination; weather condition; drone's initial pose; etc.
\subsubsection{Actors placement}
To create several scenarios from a larger scene in the City Park environment, first a Region of Interest (ROI) is selected for each autonomous landing mission in the form of a square of $30m^2$. Then, the number of actors $n_a \in [80, 120]$ within the ROI is randomly generated, as well as the initial position of the actors in the ROI, avoiding overlap between actors, all of these using uniform distributions. Also the kind of actor and their characteristics are randomly generated. The initial position of the drone in the ROI is also randomized.



\subsubsection{Actors movement}
After the initial position of all the actors is set, an algorithm selects randomly some of the actors to move during the experiment, in order to emulate the dynamic nature of scenes containing people, where the people is constantly moving at will, with dynamics unknown to the vision algorithm. A random walk algorithm in 2-Dimensions on a lattice was implemented, according to Equation \ref{eqrandomwalk}, where the lattice is represented by $\mathbb{Z}^{d}$, let $(x^a_k,y^a_k) \in \mathbb{Z}^{d}$ be the discrete position of an actor in the lattice at time $k$, then
\begin{equation}
\begin{array}{l}
x^a_k=x^a_{k-1}+0.2\alpha_x ;
\\
y^a_k=y^a_{k-1}+0.2\alpha_y
\end{array}
\label{eqrandomwalk}
\end{equation}
where $\alpha_{x,y} \in \{-1,0,1\}$ is randomly generated at each time step, and the actor position is updated $10$ times per second.


\subsection{Autonomous landing algorithm}
\begin{figure}[t]
\centerline{\includegraphics[width=0.5\textwidth]{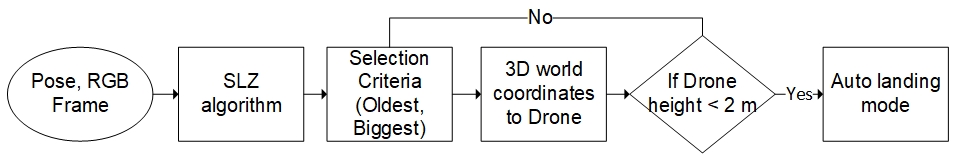}}
\caption{Flowchart of the autonomous landing algorithm based in the proposals by the SLZ algorithm.}
\label{fig:landingAlgorithm}
\end{figure}
As previously stated, AirSim is used to simulate the physics of a virtual UAV, and provides useful information of the emulated most common sensors in a drone, such as an Inertial Measurement Unit (IMU), altimeter, and different cameras. This way, integration with the virtual drone in AirSim is analogous to integration with a real drone, and allow us to test different movement policies and landing strategies. Figure \ref{fig:landingAlgorithm} represents the algorithm used for the autonomous landing task. In the first stage the Drone's pose from the simulated IMU, and a frame from a virtual down-looking camera attached to the drone are send to the SLZ algorithm described in Section \ref{sec2}. Then, a set of possible landing areas is provided in real-time by the SLZ vision algorithm, parameterized by the center of the circular SLZ in real-world $\mathbf{F_W}$ coordinates, radius an object identifier (ID) for each SLZ. The detected SLZ are updated at each iteration, and being this a mobile platform in a possibly dynamic environment (people moving) the available landing regions may be changing dynamically, resulting in one of the main challenges for this problem. Hence, it is necessary to define an adequate criteria to select a landing zone. Indeed, one of the advantages of the  proposed virtual framework is the ability to study different criteria for selecting ``the best'' SLZ, for instance, considering the size (the bigger), the distance to the drone, or the time consistency (older SLZ). After selecting a target landing zone, a 3D coordinate in $\mathbf{F_W}$ is parsed as a way-point command to the drone, $2$m above the center of the selected SLZ. If the altitude of the drone is less or equal than $2$m, a landing command is sent to the drone and the mission is over, otherwise, the drone must move towards the target SLZ and iterate the algorithm updating the image frame and the UAV pose. 


 









\section{Experimental results} \label{sec4}
\begin{figure*}[h]
\centerline{\includegraphics[width=0.8\textwidth]{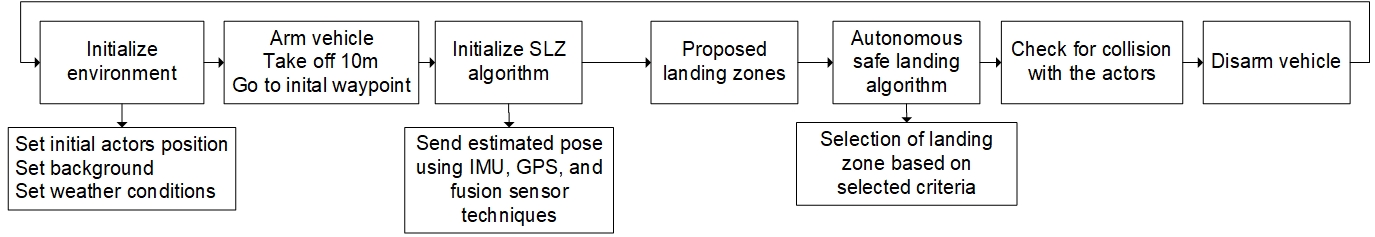}}
\caption{Flowchart of the validation process for the autonomous landing algorithm  using virtual environments.}
\label{experimentflow}
\end{figure*}

To validate the autonomous landing algorithm for UAVs in populated areas, several tests have been carried out in the virtual environment, evaluating the algorithm under a wide variety of scenarios with different conditions. For each scenario condition, $100$ iterations of autonomous landing missions have been performed. We consider four different scenario conditions: static actors, $10\%$ of the actors moving, $20\%$ of the actors moving, and dynamic weather conditions including dust, rain, snow, fog, etc. At each landing iteration, random initial conditions are generated as described in the previous section. Furthermore, two different criteria has been studied for the SLZ selection, considering the biggest SLZ available, and the oldest SLZ. Figure \ref{experimentflow} describes the steps of each autonomous landing iteration, first, the random environment is initialized, then, the UAV takes-off and moves towards a random initial position, then, the autonomous landing algorithm is started and a target landing zone is selected at each time step, checking for collisions with the human actors. Finally, different metrics are obtained, as described in the following Subsection. At the end of the $100$ runs, some statistics are obtained and a quantitatively comparison is possible. A video demonstrating the autonomous landing evaluation is available at \url{https://youtu.be/oMhk_b7QhRw}

\subsection{Evaluation metrics}
In order to measure the performance of the landing algorithm in real-time, we establish a successful landing rate as the number of missions in which the virtual UAV was able to land without colliding with the human actors. This is the most important metric in our study, providing a direct measure of the number of accidents prevented by the algorithm.

Just consider the rate of successful landings does not provide a complete picture of the performance of the SLZ algorithm and the criteria, hence, other metrics are proposed. ``Warning'' represents the percentage of times some actor ended within a safety radius, either $1$m or $1.5$m, around the landing point. Moreover, the SLZ area is important given that larger landing zones offer increased safety to the people around in case of errors, allowing aware people to move to avoid the collision. The ``nearest person'' measures the average euclidean distance between the drone and the closest person during landing. Finally, the IoU is computed for the target landing zone against the closest ground truth people-free region.



\subsection{Results}
Table \ref{tab:results} presents the obtained results over $100$ iterations for each case of study. To establish a comparison baseline, a random landing was included in this experiment considering only static actors, emulating the scenario where a drone must perform a blind emergency landing, since not all the ROI was occupied by actors, there is a possibility that some of the landings are successful. In this case, the autonomous landing reached a $59\%$ of successful landings. Firstly, we evaluate both criteria, bigger SLZ and older SLZ, for a static environments. As expected, for the case of an static scenario both criteria performed fairly well, reaching $97\%$ and $99\%$ respectively. Due to computational limitations, we only tested the strategies for movement in the $10\%$ and $20\%$ of the actors in the scene, nevertheless, this is enough to analyze the behavior of the landing algorithm under dynamic conditions. Adding mobile actors in the the scene considerably increases the difficulty of the task, resulting in a decrease in the performance of the strategies. Although there is still room for improvement, both strategies obtained acceptable results when compared with the random landing in the easier static scenario, suggesting that the use of this kind of autonomous landing protocols in case of an emergency landing could considerably reduce the risk of injuring people. More in particular, the obtained results suggest that the use of the older SLZ is better suited to account for people moving with unknown dynamics, given that the older SLZ tends to be in regions where people are less likely to be, which is in compliance with the results in \cite{gonzalez-trejo21_visual_based_safe_landin_uavs_popul_areas}. Also, the proposed pipeline was tested under dynamic weather conditions, such as dust, mist, rain, snow, casting similar results, proving that the algorithm is robust to noisy conditions. With regards to SLZ area, the oldest SLZ strategy tended to decrease around $30\%$ compared with the biggest SLZ strategy, in each of the scenario and strategy combinations, suggesting that there is a trade-off between being robust against people moving, and providing additional safety by choosing a larger SLZ, and probably a multi-objective cost function would be a good alternative selection criteria.

\begin{table*}[h]    
\centering
\caption{Performance comparison for Autonomous Landing under different conditions.}
\centering
\begin{tabular}{|c|c|c|c|c|c|c|}
\hline
Scenario conditions / Strategy & \multicolumn{1}{|c|}{Successful \%} & \multicolumn{1}{|c|}{Warning 1 [m]} &  
\multicolumn{1}{|c|}{Warning 1.5 [m]} &  
SLZ area [$m^2$] & \multicolumn{1}{|c|}{Avg. Nearest Person [m]} & IoU \\
\hline
Static / Random landing & 59\% & 2.7\%  & 10\% & * & 1.1579 & *\\
\hline
Static / Oldest SLZ & 97\% & 0.0\% & 0.4\% & 6.153 & 2.2197 & 0.484 \\
\hline
Static / Biggest SLZ & 99\% & 0.0\% & 0.3\% & 9.621& 2.4508 & 0.593 \\
\hline
Moving actors 10\% / Oldest SLZ & 87\% & 0.06\% & 0.16\% & 4.5996  & 1.9706 & 0.394\\
\hline
Moving actors 10\% / Biggest SLZ & 80\% & 0.01\% & 0.1\% & 9.079 & 1.7087 & 0.607\\
\hline
Moving actors 20\% / Oldest SLZ & 84\% & 0.06\% & 0.14\% & 4.374& 1.5848 & 0.421\\
\hline
Moving actors 20\% / Biggest SLZ & 76\% & 0.09\% & 0.2\% & 8.042& 1.5136 & 0.549\\
\hline
Dynamic Weather / Oldest SLZ & 86\% & 0.08\% & 0.16\% & 4.495 & 1.8451 & 0.527 \\
\hline
Dynamic Weather / Biggest SLZ & 80\% & 0.06\% & 0.1\% & 6.9582 & 2.074 & 0.540\\
\hline
\end{tabular}
\label{tab:results}
\end{table*}

\section{Conclusions and future work} \label{sec5}
Autonomous landing of UAVs in populated environments is a new and exiting research axis, with a lot of challenges and great potential in real-world applications, in fact, providing UAVs with this capabilities is a key aspect to explode the full potential of drones in urban environments, particularly in emergency landing situations. However, due risk involved, it is of crucial importance to fail-proof such autonomous landing algorithms beforehand, in order to fine-tune the algorithms and the select the best suited one, and more important, to guarantee the safety of the people around, which turns out to be a very difficult task. 

In this work, we have proposed an evaluation framework using virtual environments for safely and thoroughly testing the autonomous landing algorithms in populated areas. The virtual environments were generated with the Unreal graphics engine, whereas a virtual drone is simulated using AirSim. This framework allows us to iterate the autonomous landing experiment more than $900$ times, under different randomized conditions, reducing time and resources needed for the testing.  

Also, two variations of a vision-based safe landing zones detector were evaluated in real-time, either selecting the biggest available landing zone or the oldest available one. Different metrics were computed and a quantitatively comparison study was provided. The study suggests that the use of the autonomous landing algorithms in populated environments may considerably reduce the risk of injuring people. Furthermore, it was found that the oldest safe landing zone tends to be more robust in the case of a dynamic environment where people is moving in the scene, obtaining up to $86\%$ of success rate in the most difficult case of study, with a $20\%$ of the actors moving. Accordingly, a good alternative approach would be to combine both criteria in the form of a cost function.

Although there is still room for improvement, the obtained results are fairly promising and may help to reduce the risk of injuring people during civilian applications in urban areas, specially taking into account the level of difficulty and the safety constrains involved in the task.

\bibliographystyle{IEEEtran}
\bibliography{IEEEabrv,references}

\end{document}